# Research on visual simultaneous localization and mapping technology based on near infrared light

Rui Ma#, Mengfang Liu#, Boliang Li, Xinghui Li*
*Tsinghua Shenzhen International Graduate School, Tsinghua University, Shenzhen 518055, China*

**Abstract:** In view of the problems that visual simultaneous localization and mapping (VSLAM) are susceptible to environmental light interference and luminosity inconsistency, the visual simultaneous localization and mapping technology based on near infrared perception (NIR-VSLAM) is proposed. In order to avoid ambient light interference, the near infrared light is innovatively selected as the light source. The luminosity parameter estimation of error energy function, halo factor and exposure time and the light source irradiance correction method are proposed in this paper, which greatly improves the positioning accuracy of Direct Sparse Odometry (DSO). The feasibility of the proposed method in four large scenes is verified, which provides the reference for visual positioning in automatic driving and mobile robot.

**Highlight:**
(1) Near-infrared light is used as the lighting source to avoid ambient light interference and improve the imaging quality of VSLAM system.
(2) The photometric parameter estimation and photometric correction of the image are proposed to improve the pose estimation positioning accuracy of the VSLAM system.
(3) Near-infrared images of four large scenes of indoor, outdoor, underground parking lot and pipeline tunnel were collected to verify the effectiveness of the proposed VSLAM system.

## 1. Introduction

The Simultaneous Localization and Mapping (SLAM) technology is used to navigate and build maps in unknown environments through cameras, lidar and other sensors, and it has extensive applications in robotics, vehicle autonomous driving, drones, augmented reality and other fields[1].

SLAM originated from the probabilistic SLAM problem at the IEEE Robot and Automation Conference held in San Francisco in 1986[2], and experienced three stages of initial theoretical exploration (1986-2004), algorithmic framework development (2004-2015), and system robustness improvement (2015-now)[3]. According to the sensor classification, the SLAM technology can be divided into laser SLAM, visual SLAM , and multi-sensor fusion SLAM. Laser SLAM is scanned by lidar, who are suitable for indoor environment but inaccurate positioning in a single repeated environment[4-6]. Visual SLAM captures images through the camera, acquires positions and maps through image pixels and features, and is suitable for textured rich scenes. In addition, visual SLAM has the advantages of low cost and small size, which can provide intuitive visual input[7-9]. Multi-sensor fusion SLAM combines the laser, camera, inertial measurement unit (IMU) together to collect data, which can improve the robustness and positioning accuracy of the system.

Visual SLAM can be divided into three series, monocular, multiocular and RGB-D camera according to the type of cameras, and it can also be divided into direct, semi-direct and indirect algorithms according to the image information processing method[10-13]. The direct algorithm does not rely on the extraction and matching of feature points, but directly uses the pixel gray scale information of the image to estimate the motion of the camera and the 3 D structure of the scene, avoiding the time consumption of feature point extraction and matching, and is very effective in the dense and semi-dense map construction[14-16]. The indirect method, namely the feature point method, first extracts the feature points such as angles and edges from the image, and then calculates the descriptors of the feature points, which performs well in the texture-rich environment and can provide a more accurate camera motion estimation. Its representative algorithm, ORB-SLAM (oriented FAST and rotated BRIEF)[17], was proposed in 2015, which supports monocular, binocular and RGB-D cameras with good versatility. However, the calculation of feature points and descriptors of indirect method is very time-consuming, which is easy to fail in the weak texture environment, showing great limitations. It is not suitable for the platform with high real-time demand and low computing power. The semi-direct method combines the rapidity of the direct method and the accuracy of the indirect method. First, the semi-direct method quickly estimates the camera in the approximate position, and then uses the feature point method to refine the projection point processing. However, the algorithm is relatively complex and difficult to realize. In 2014, Forster et al. proposed Semi-Direct Monocular Visual Odometry SLAM (SVO-SLAM) [18], which combines direct and indirect methods, first tracking some key angles similar to indirect method, and then estimates the motion and position of the camera according to the information around the key angles. The calculation speed is very fast, but with relatively low accuracy. In conclusion, the direct method of coupled data association and pose estimation is more suitable for the positioning of continuous images, and has more advantages in algorithm implementation and deployment.

In the last 10 years, critical algorithms and classical frameworks of the visual SLAM direct method have been emerging. In 2011, Newcombe et al. proposed DTAM[19], which is a high-resolution and intensive 3-D model of scenes generated in real time by image processing units. In 2014 Engel et al proposed LSD-SLAM[20], which can track the camera motion by directly matching the gray values of the image block and uses a probabilistic model to represent the semi-dense depth map. Engel et al of the Computer Vision Laboratory of the Technical University of Munich proposed DSO-SLAM (Direct Sparse Odometry, DSO) in 2017[21], which outperformed LSD-SLAM in terms of accuracy, stability and speed. DSO enhances the robustness of the direct method by photometric calibration, and enables real-time performance on an ordinary CPU.

VSLAM performance is affected much by light in the actual scene, showing the problem of poor imaging quality and insufficient system robustness under degradation light. Near infrared light is friendly to human eyes and has good penetration ability, by which imaging can still have high resolution and contrast even under poor environmental conditions such as cloudy, rainy and foggy days. Since there is visible light in the environment and almost no near infrared light, the selection of near infrared light source can eliminate the ambient light interference and improve the imaging quality.

In view of the problems of visual synchronous positioning and mapping (VSLAM) susceptible to environmental light interference and luminosity inconsistency, this paper proposes the visual synchronous positioning and mapping technology based on near infrared perception. The main contributions of this article are provided as follows:

(1) Near-infrared is used as the lighting source, so as to avoid ambient light interference and improve the imaging quality of VSLAM system.

(2) The photometric parameter estimation and photometric correction of the image, so as to improve the positioning accuracy of the VSLAM system pose estimation.

(3) Near-infrared images of four large scenes of indoor, outdoor, underground parking lot and

pipeline tunnel were collected to verify the effectiveness of the proposed VSLAM system.

## 2. The VSLAM system based on NIR

### 2.1 The VSLAM system and the workflow

The VSLAM system consists of two parts: scene data acquisition and scene data processing, and its system design is shown in Fig.1. VSLAM The core components of the system include near-infrared light source, binocular camera, Leica full station, Leica prism and so on. Near-infrared light refers to the electromagnetic wave with a wavelength of 780nm~2500nm. As the sensitivity of the camera gradually decreases with the increase of wavelength, the near-infrared light source with a wavelength of 850nm is selected. A narrow-band filter with a central wavelength of 850nm was installed in front of the binocular camera to eliminate ambient light interference and improve the imaging quality. The near-infrared light reflected by the Leica prism is received by the Leica full station to estimate the trajectory and pose of the camera for positioning and mapping.

The pose estimation platform includes near-infrared light source, binocular camera, Leica full station, Leica prism, voltage converter, computer, power supply, etc. The pose estimation and trajectory tracking of real scenes are done on the site pose estimation platform.

Scene data processing is mainly to make the luminosity correction of the data collected by the VSLAM system through the camera. The SLAM algorithm evaluates the pixel of the collected image, which is greatly affected by the light source. In the process of moving the camera and the light source, the irradiation Angle and distance of the light source will change, and the irradiance of the light source in the scene will also change. Therefore, it is necessary to accurately model the irradiance of the light source model, estimate the luminosity parameter and correct the image.

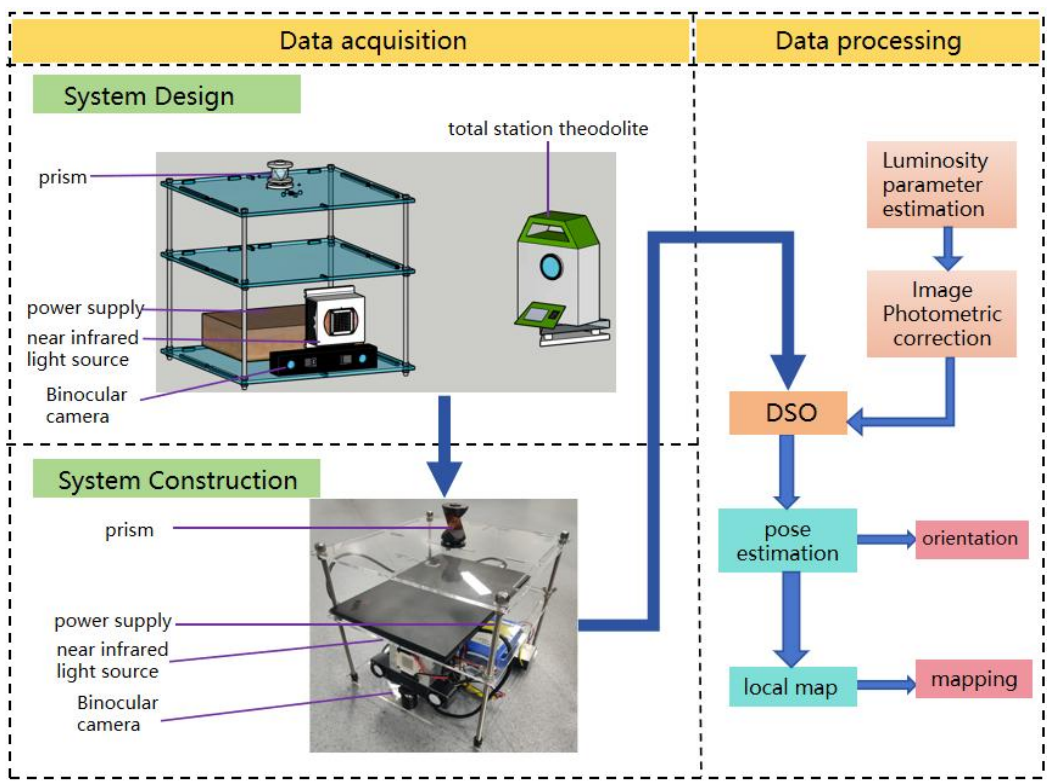

Fig.1 The system of VSLAM

### 2.2 NIR sensing imaging

In the process of pose estimation and local mapping of VSLAM system, the imaging quality directly affects the extraction effect and matching accuracy of the image field attractions.

Compared with visible light, near-infrared light has good penetration ability and anti-environmental light interference ability, and the clarity and contrast of imaging are not affected by clouds, rain, fog and other weather, which has good robustness. Figure 2 compares the visible and near-infrared light images taken in the on and off states, respectively. As can be seen from Fig. 2, the near infrared image is almost unaffected by the indoor environmental light source, and the difference between the image before and after the opening of the near infrared light source is relatively small, which confirms the robustness of the near infrared perception imaging.

|  |  | Visible light | | | Near infrared light | | |
|---|---|---|---|---|---|---|---|
|  |  | ON | OFF | Difference | ON | OFF | Difference |
| Near infrared light | OFF |  |  |  |  |  |  |
|  | ON |  |  |  |  |  |  |

Fig.2 The comparison of images under visible and near infrared light

The information entropy of images is an important indicator of local illumination information and one of the evaluation indicators of image imaging quality. Information entropy refers to the bit average of the gray scale set of images, which describes the average information amount of the image source and can be used to evaluate the complexity of the image, such as contrast, detail, texture and other features. The information entropy is calculated as follows:

$$H(X)=-\sum_{i=1}^{n} P(x_i) \log_b P(x_i) \qquad (2\text{-}1)$$

Where H(X) is the information entropy of the image pixel; $n$ is the number of image pixels, $n=255$; $P(x_i)$ is the probability that the image pixel $x$ takes the $i$-th state; information entropy $b=1$, because the unit of information is bit.

In order to study the influence of illumination intensity on imaging quality, NIR images and visible images were collected in the dark room and outdoor light, and the imaging effect is shown in Fig. 3. With the increase of NIR illumination intensity, the sharpness and resolution of the image become higher.

|  |  | 0W | 20W | 40W | 50W | 70W | 100W | Visible Light |
|---|---|---|---|---|---|---|---|---|
| Dark Room | senerio 1 |  |  |  |  |  |  |  |
|  | senerio 2 |  |  |  |  |  |  |  |
| Night outdoors | senerio 3 |  |  |  |  |  |  |  |
|  | senerio 4 |  |  |  |  |  |  |  |

Fig.3 The images of different illumination intensity

To further explore the imaging quality in different scenes, the information entropy of the image frames in the dark room and outside the night room was collected in Fig. 3, as shown in

Table 1. The highest information entropy of the dark room and outdoor sequence at night is 7.33 and 7.16, respectively, which are higher than the 7.16 and 6.98 of the visible light, indicating that NIR imaging contains richer texture and details. This experiment provides a reference for the selection of near-infrared lighting power under different scenarios and lighting conditions. For indoor scenes, the lighting power of 50W-100W can be selected according to the space area; for outdoor scenes, the lighting power of over 100W can be selected.

Table 1 Information entropy of NIR and visible light images at different illumination intensities

| scene | Image frame | Near infrared lighting | | | | | | visible light |
|---|---|---|---|---|---|---|---|---|
| | | 0W | 20W | 40W | 50W | 70W | 100W | |
| Dark room | Frame 1 | 2.93 | 4.89 | 5.72 | 6.35 | 7.38 | 7.36 | 7.01 |
| | Frame 2 | 1.99 | 4.25 | 5.42 | 6.37 | 7.33 | 7.30 | 7.25 |
| | Sequence average | 2.68 | 4.41 | 5.32 | 6.38 | 7.33 | 7.29 | 7.16 |
| Outdoor at night | Frame 3 | 3.52 | 3.84 | 4.77 | 4.44 | 6.93 | 7.32 | 7.14 |
| | Frame 4 | 2.59 | 3.25 | 4.06 | 3.92 | 6.35 | 7.12 | 6.83 |
| | Sequence average | 2.87 | 3.81 | 4.56 | 4.56 | 6.83 | 7.16 | 6.98 |

In VSLAM, feature matching performance is a key indicator of localization accuracy. The ambient illumination change has some influence on the imaging quality of NIR images, and also affects the feature point matching and localization accuracy in the SLAM system. Fig. 4 studies the feature matching of visible light images and near-infrared images in the same indoor environment. Compared with visible light, the near-infrared image features a larger number of feature points and a more uniform distribution, especially covering the edge area of the image that contains important geometric information. In conclusion, the near infrared light source lighting in VSLAM system has great advantages.

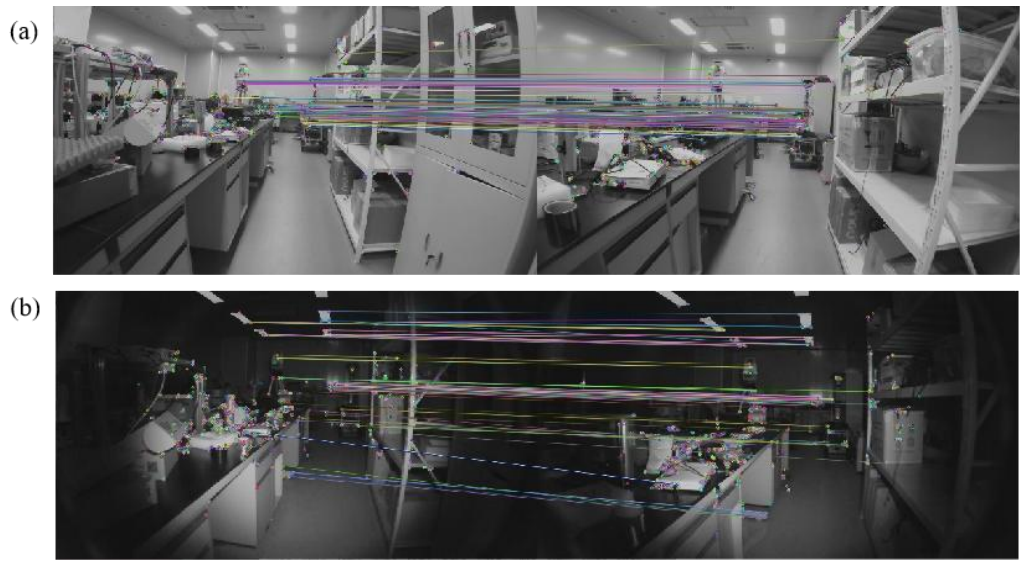

Fig.4 The feature point match images (a)Visible light (b) near infrared light

**2.3 Photometry correction method**

The direct method of pose estimation in the VSLAM system is based on the luminosity constant assumption, assuming that the irradiance of the same light source in the scene does not change when the camera perspective changes. The direct method uses the minimum error to calculate the irradiance difference of the camera for the object in the scene. However, in the actual operation of VSLAM system, with the movement of the camera and near infrared light source, the intensity and Angle of the light source on the surface of the scene object change; the change of illumination intensity directly affects the image gray value and the pixel intensity value of the field affects the shadow on the surface of the object, thus affecting the visual appearance of the object. In view of the problem of inconsistent irradiance at different times in the same scene, this paper proposes the method of luminosity correction. By estimating photometric parameters such as error energy function, halo factor and exposure time, the irradiance of image source is corrected, so as to maintain the luminosity consistency as much as possible when the illumination changes.

According to the literature[1], the error energy function of the luminosity was established. Given a set of field points $P$ tracked on a series of images, where the point $p \in P$ is the observation point in the frame $F_p$, the error energy function is expressed as:

$$E = \sum_{p \in P} \sum_{i \in F_p} \left\| \frac{f^{-1}(I_i(u_p))}{e_i V(r_i^p)} - L^p(d_p) \right\|_h \tag{2-2}$$

Where $I_i(u_p)$ is the output intensity of $p$ in the image $i$, namely the image brightness value. $e_i$ is the exposure time of the image $i$, $L^p$ is the radiance of the point $p$, $f(*)$ is the estimated camera response function, $V(r_i^p)$ is the halo factor. This paper uses the Huber norm, $||.||_h$, used for robust estimation, parametrized as $h \in R$. And the energy formula assumes that each field attraction is located on a Lamber surface.

The residues $r$ of the error energy function are optimized by an analytic Jacobian matrix, as shown in Eq. (2-3) - (2-6). The parameters were gradually adjusted by an iterative approach against the information provided by the Jacobian matrix to minimize the residual difference $r$. This approach not only improves the accuracy of the parameter estimation, but also improves the efficiency of the entire optimization process.

$$\mathbf{J} = \left( \frac{\partial r}{\partial \mathbf{c}}, \frac{\partial r}{\partial \mathbf{v}}, \frac{\partial r}{\partial e_i} \right) \tag{2-3}$$

$$\left( \mathbf{J}^T \mathbf{W} \mathbf{J} + \lambda \, \mathrm{diag}\left( \mathbf{J}^T \mathbf{W} \mathbf{J} \right) \right) \Delta \mathbf{x} = -\mathbf{J}^T \mathbf{W} \mathbf{r} \tag{2-4}$$

$$w_r^{(2)} = \frac{\mu}{\mu + \left\| \nabla F_i(x_i^p) \right\|_2^2} \tag{2-5}$$

$$J^p = \left( \frac{\partial r}{\partial L_P} \right) \tag{2-6}$$

Eq. (2-3) represent the complete Jacobian matrix, $c = (c_1, c_2, c_3, c_4)$ and $v = (v_1, v_2, v_3)$. Let $e$ represents the vector of all exposure times. The state update of the parameters $\Delta x = (\Delta c, \Delta v, \Delta e)$ can be obtained by solving the normal equation. Eq. (2-4) $diag(A)$ shows that extracting diagonal elements from the input matrix $A$ and forming a new diagonal matrix. $r$ is a summary vector of residues generated from all individual observations and $\lambda$ is the damping factor used in the optimization algorithm. $W$ is a diagonal weight matrix, which is constructed by combining the sum of the two weighting factors $W_r^{(1)}$ and $W_r^{(2)}$. $W_r^{(1)}$ is for robust Huber estimation, $W_r^{(2)}$ is set to prevent high gradient positions from overaffecting the overall optimization and is used to weight the residuals of the corresponding positions. In Eq. (2-5), there $\mu \in R^+$ is a constant greater than zero, $||\nabla Fi(x_i^p)||_2^2$ is the square of the corresponding position $x_i^p$ of the image $Fi$, that is, the $i-th$ position gradient $L2$ norm. The final weight value of the residue is given by multiplying the two weights $W_r^{(1)}$ and $W_r^{(2)}$ together. Eq. (2-6) is the partial derivative of the residual difference relative to the irradiance of the scene.

In the process of residual $r$ optimization by Jacobian determinant, the light source and irradiance are constantly updated, and the update expression is as follows:

$$\Delta L = \frac{-J_p^T W_p r_p}{(1+\lambda)J_p^T W_p J_p} \quad (2\text{-}7)$$

Where, $W_p$ representing the corresponding point $p$ optimization weight, $r_p$ representing the superimposed residue, $\lambda$ is the damping factor used in the optimization algorithm.

In order to verify the accuracy of the proposed method for the estimation of camera luminosity parameters, the results of the estimation of camera luminosity parameters were experimentally verified on the public data set TUM Mono VO[22], and the experimental results were shown in Fig.5.

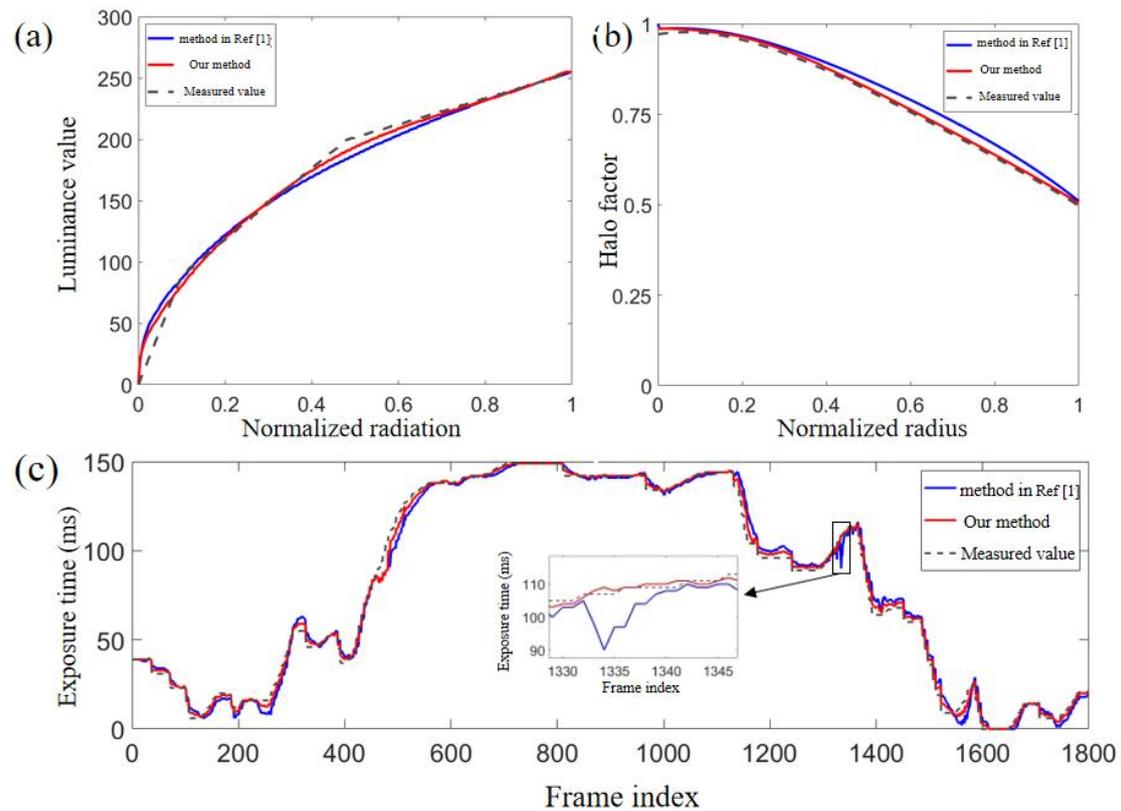

Fig. 5 Photometry parameters estimation Experimental results
(a) response function; (b) halo factor; (c) exposure time

The blue solid line, red solid line and black dashed line represent the comparison method[1],

the proposed method in this paper and the actual true value respectively. By analyzing the experimental data, the average error of the comparison method is 5.76 and the maximum error is 27. The average error of this method is 3.56, and the maximum error is 22. For the estimation of camera halo coefficient, the average error of the comparison method is 0.02, and the maximum error is 0.04. The average error of this method is 0.01, and the maximum error is 0.02. For the estimation of camera exposure time, the average error of the comparison method is 3.41ms, and the maximum error is 31ms. The average error of this method is 1.72ms, and the maximum error is 14ms. The above estimation results of the three parameters show that the proposed method achieves a more accurate estimation of camera luminosity parameters.

## 3. Experiments
### 3.1 Data acquisition

In this paper, a pose estimation platform for the VSLAM system as shown in Figure 1. The hardware system uses small binocular camera D1000-IR-120, 360° Leica tracking prism, Leica TS60 full station, voltage conversion module, 24V-850nm near-infrared light source, computer, etc. The binocular camera has a six-axis IMU sensor and an infrared active detector, with an 850mm narrow-band filter in front of the camera lens. The software system is based on Linux operating system Ubuntu 18.04 and ROS Melodic version, and the third-party libraries configured mainly include Opencv 3.4.3, Ceres1.14.0, Pangolin0.6, Eigen-3.3.4, etc.

Four different types of representative large scenes of indoor, outdoor, underground parking lot and tunnel and tunnel were selected for near-infrared image data sequence collection, as shown in Fig. 6. The acquired NIR images are shown in Fig. 7. The four scenes have their respective illumination characteristics and imaging conditions. The indoor environment light is relatively stable, but will be affected by the natural light sources through the window; the outdoor scene light conditions are complex and changeable, affected by the sunshine height, cloud and fog weather; most of the light sources in the underground parking lot belongs to artificial lighting source, the ground is mostly reflective, the light conditions may change with the vehicle; the tunnel is a closed environment, mostly artificial lighting light source, the tunnel is made of concrete, reflective performance is poor, and the light distribution is uneven.

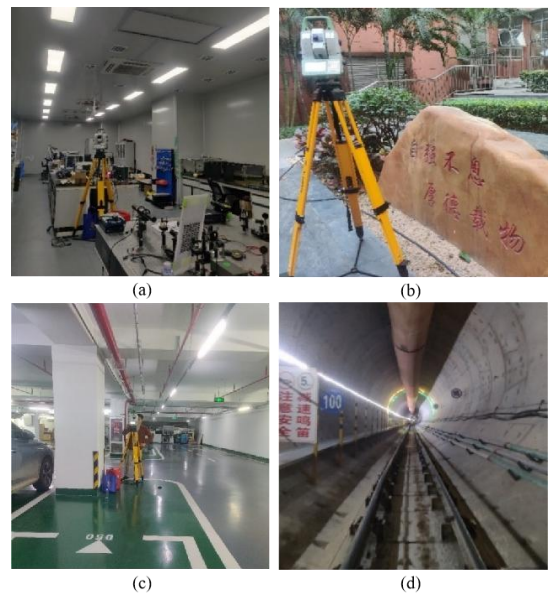

Fig. 6 NIR images were collected in large scenarios by the VSLAM system
(A) indoor room; (b) outdoor scene; (c) underground parking lot; (d) pipe tunnel

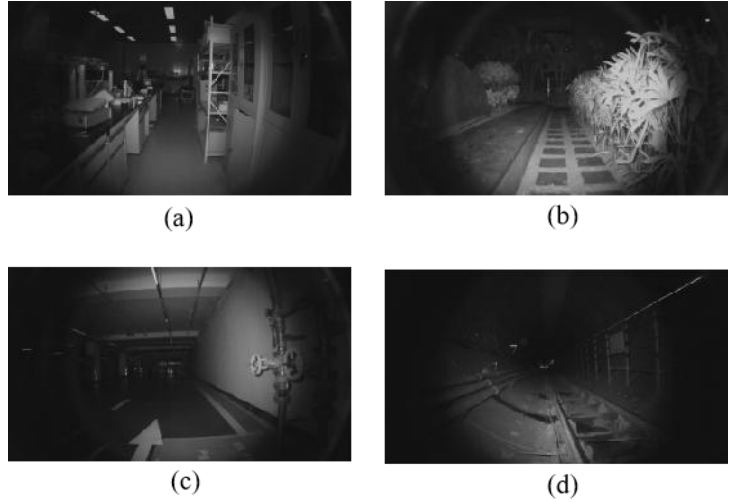

Fig. 7 NIR images collected in large scenes

(A) indoor room; (b) outdoor scene; (c) underground parking lot; (d) pipe tunnel

In the trajectory tracking experiment, the pose estimation platform moves with a certain trajectory and captures the light reflected by the Leica prism is captured through the Leica full station. In the four scenes of indoor, outdoor, underground parking lot, pipe and tunnel, when the tracking line of sight is not blocked, complex and representative movement tracks are collected as far as possible. The data acquisition frequency in the VSLAM system is set to a time interval of 0.1s or an interdistance interval of 2mm to trigger the acquisition mode simultaneously, thus improving the validity of the data. Fig. 8 shows the movement trajectory of the partially collected data series.

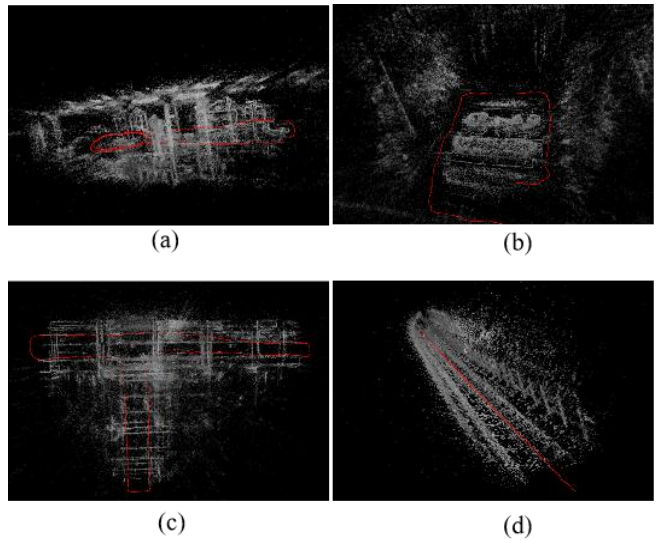

Fig. 8 Movement trajectory in the large scene

(A) indoor room; (b) outdoor scene; (c) underground parking lot; (d) pipe tunnel

**3.2 Data processing**

The proposed method in this paper adds a luminosity correction module to the direct method DSO, and the images collected in the four scenes are estimated by DSO, improved DSO[1] and the proposed method to calculate the absolute trajectory error, as shown in Table 2. The proposed photometric correction method has a positive effect on the direct method of DSO positioning accuracy. In four large scenarios: indoor, outdoor, underground parking lot and pipe-to-tunnel, the proposed method improves by 57.6% on average than the DSO method. Fig. 9 shows the two-dimensional trajectory error heat map of the DSO method and the proposed method under

50W near-infrared illumination. It can be seen from Figure 9 that compared with DSO method, the actual motion trajectory and pose estimation trajectory are more compatible, showing that the proposed method can estimate the position more accurately and has more advantages in terms of positioning accuracy and robustness.

Table 2 Absolute trajectory error results based on large scene data method (unit: m)

| data series | DSO | Improved DSO[1] | Ours | Increase the percentage |
| --- | --- | --- | --- | --- |
| Room_20 | 0.782 | 0.358 | 0.333 | **7.5%** |
| Room_50 | 0.875 | 0.137 | 0.123 | **11.4%** |
| Room_100 | 1.023 | 0.435 | 0.398 | **9.3%** |
| Outside_20 | 0.291 | 0.171 | 0.150 | **14.0%** |
| Outside_50 | 0.372 | 0.283 | 0.254 | **11.4%** |
| Outside_100 | 0.951 | 0.375 | 0.342 | **9.6%** |
| Parking_20 | 0.391 | 0.185 | 0.165 | **12.1%** |
| Parking_50 | 0.473 | 0.183 | 0.167 | **9.6%** |
| Parking_100 | 0.386 | 0.141 | 0.128 | **10.2%** |
| Tunnel_20 | 1.101 | 0.563 | 0.527 | **6.8%** |
| Tunnel_50 | 1.038 | 0.542 | 0.503 | **7.8%** |
| Tunnel_100 | 1.081 | 0.586 | 0.544 | **7.7%** |

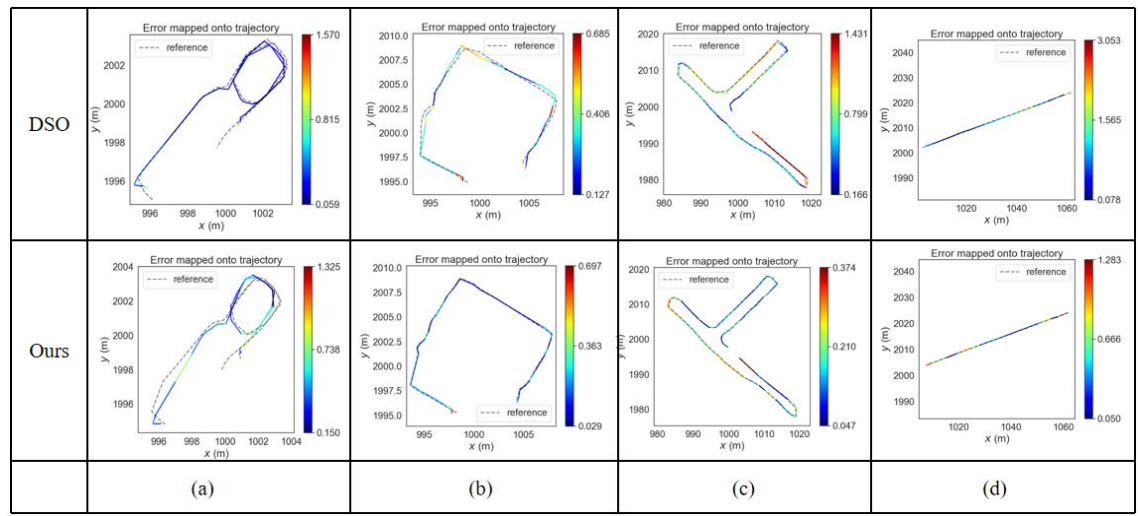

Fig. 9 The proposed method and DSO 2 D trajectory error heat map in large scenes of 150 W near-infrared illumination

(a) Indoor room; (b) outdoor scene; (c) underground parking lot; (d) pipe tunnel tunnel

## 4. Conclusion

This paper presents a visual synchronous positioning and mapping technology based on NIR perception (NIR-VSLAM), The interference of the ambient light is avoided by the near-infrared light source illumination, It provides high-quality scene images for the subsequent pose estimation; By estimate photometric parameters such as error energy function, gradient halo factor and exposure time, To minimize the residual difference, Thus, correcting for the light source irradiance of the scene, The proposed photometric correction method is combined with the pose estimation algorithm DSO, It greatly improves the positioning accuracy of VSLAM system; The proposed VSLAM in indoor, outdoor, underground parking lot and pipe tunnel four large scenarios and verify the feasibility and effectiveness, It provides an application reference for visual positioning in autonomous driving and mobile robots.